\documentclass[sigconf]{acmart}
\usepackage{enumitem}
\usepackage{circledsteps} 
\usepackage{graphicx}
\usepackage{subcaption}
\usepackage{multirow}
\usepackage{booktabs}
\usepackage{lipsum}
\usepackage{nicematrix}
\usepackage{makecell}
\usepackage{wrapfig}
\usepackage{graphicx}
\usepackage{listings}
\usepackage{xcolor}
\usepackage{multicol}
\usepackage{enumitem}
\AtBeginDocument{%
  }

\copyrightyear{2026}
\acmYear{2026}
\setcopyright{cc}
\setcctype{by-nc-nd}
\acmConference[KDD '26]{Proceedings of the 32nd ACM SIGKDD Conference on Knowledge Discovery and Data Mining V.2}{August 09--13, 2026}{Jeju Island, Republic of Korea}
\acmBooktitle{Proceedings of the 32nd ACM SIGKDD Conference on Knowledge Discovery and Data Mining V.2 (KDD '26), August 09--13, 2026, Jeju Island, Republic of Korea}
\acmDOI{10.1145/3770855.3817468}
\acmISBN{979-8-4007-2259-2/2026/08}

\begin{document}

\title{FinTexTS: Financial Text-Paired Time-Series Dataset via Semantic-Based and Multi-Level Pairing}





\author{Jaehoon Lee}\authornote{Both authors contributed equally to this research.}
\affiliation{%
  \institution{LG AI Research}
  \city{Seoul}
  \country{Republic of Korea}
}
\email{jaehoon.lee@lgresearch.ai}

\author{Suhwan Park}\authornotemark[1]\authornote{This work was done during an internship at LG AI Research.}
\affiliation{%
  \institution{Ulsan National Institute of Science and Technology}
  \city{Ulsan}
  \country{Republic of Korea}
}
\email{suhwan@unist.ac.kr}

\author{Taeyoon Lim}
\affiliation{%
  \institution{LG AI Research}
  \city{Seoul}
  \country{Republic of Korea}
}
\email{tylim@lgresearch.ai}

\author{Seunghan Lee}
\affiliation{%
  \institution{LG AI Research}
  \city{Seoul}
  \country{Republic of Korea}
}
\email{seunghan.lee@lgresearch.ai}

\author{Jun Seo}
\affiliation{%
  \institution{LG AI Research}
  \city{Seoul}
  \country{Republic of Korea}
}
\email{jun.seo@lgresearch.ai}

\author{Dongwan Kang}
\affiliation{%
  \institution{LG AI Research}
  \city{Seoul}
  \country{Republic of Korea}
}
\email{evan.kang@lgresearch.ai}

\author{Hwanil Choi}
\affiliation{%
  \institution{LG AI Research}
  \city{Seoul}
  \country{Republic of Korea}
}
\email{hwanil.choi@lgresearch.ai}

\author{Minjae Kim}
\affiliation{%
  \institution{LG AI Research}
  \city{Seoul}
  \country{Republic of Korea}
}
\email{m.j.kim@lgresearch.ai}

\author{Sungdong Yoo}
\affiliation{%
  \institution{LG AI Research}
  \city{Seoul}
  \country{Republic of Korea}
}
\email{yoosd424@lgresearch.ai}

\author{Soonyoung Lee}
\affiliation{%
  \institution{LG AI Research}
  \city{Seoul}
  \country{Republic of Korea}
}
\email{soonyoung.lee@lgresearch.ai}

\author{Yongjae Lee}
\affiliation{%
  \institution{Ulsan National Institute of Science and Technology}
  \city{Ulsan}
  \country{Republic of Korea}
}
\email{yongjaelee@unist.ac.kr}

\author{Wonbin Ahn}\authornote{Corresponding author.}
\affiliation{%
  \institution{LG AI Research}
  \city{Seoul}
  \country{Republic of Korea}
}
\email{wonbin.ahn@lgresearch.ai}

\renewcommand{\shortauthors}{Lee et al.}

\begin{abstract}
The financial domain involves a variety of important time-series problems. 
Recently, time-series analysis methods that jointly leverage textual and numerical information have gained increasing attention.
Accordingly, numerous efforts have been made to construct text-paired time-series datasets in the financial domain. However, financial markets are characterized by complex interdependencies, in which a company's stock price is influenced not only by company-specific events but also by events in other companies and broader macroeconomic factors. Existing approaches that pair text with financial time-series data based on simple keyword matching often fail to capture such complex relationships. 
To address this limitation, we propose a semantic-based and multi-level pairing framework. 
Specifically, we extract company-specific context for the target company from SEC filings and apply an embedding-based matching mechanism to retrieve semantically relevant news articles based on this context.
Furthermore, we classify news articles into four levels (macro-level, sector-level, related company-level, and target company-level) using large language models (LLMs), enabling multi-level pairing of news articles with the target company.
Applying this framework to publicly-available news datasets, we construct \textbf{FinTexTS}, a new large-scale text-paired stock price dataset.
Experimental results on \textbf{FinTexTS} demonstrate the effectiveness of our semantic-based and multi-level pairing strategy in stock price forecasting. 
In addition to publicly-available news underlying \textbf{FinTexTS}, we show that applying our method to proprietary yet carefully curated news sources leads to higher-quality paired data and improved stock price forecasting performance.

\end{abstract}


\received{8 February 2026}
\received[accepted]{16 May 2026}

\maketitle

\section{Introduction}
In the financial domain, effectively modeling time-series data 
has long been considered critically important, 
as many economic signals, including stock prices, exchange rates, commodity prices, and market volatility, evolve over time. 
Motivated by this importance, numerous deep learning-based time-series models have been proposed to address financial time-series problems~\cite{luo2018neural, nakagawa2019deep, zhu2025fincast}. Among various time-series tasks, stock price forecasting has received particularly strong research attention~\cite{li2024master, yoo2021accurate}.

Recently, the great success of vision–language multimodal models~\cite{radford2021clip, li2022blip, liu2023llava} has sparked growing interest in multimodal learning across a wide range of domains. Accordingly, the development of text-time series (text-TS) multimodal models has also attracted increasing attention. By jointly leveraging numerical time-series data and textual information such as news articles, these text-TS models offer greater potential than traditional approaches solely relying on numerical time-series data~\cite{liu2024timemmd, seo2025air}.

\begin{figure*}[t]
    \centering
    \includegraphics[width=\linewidth]{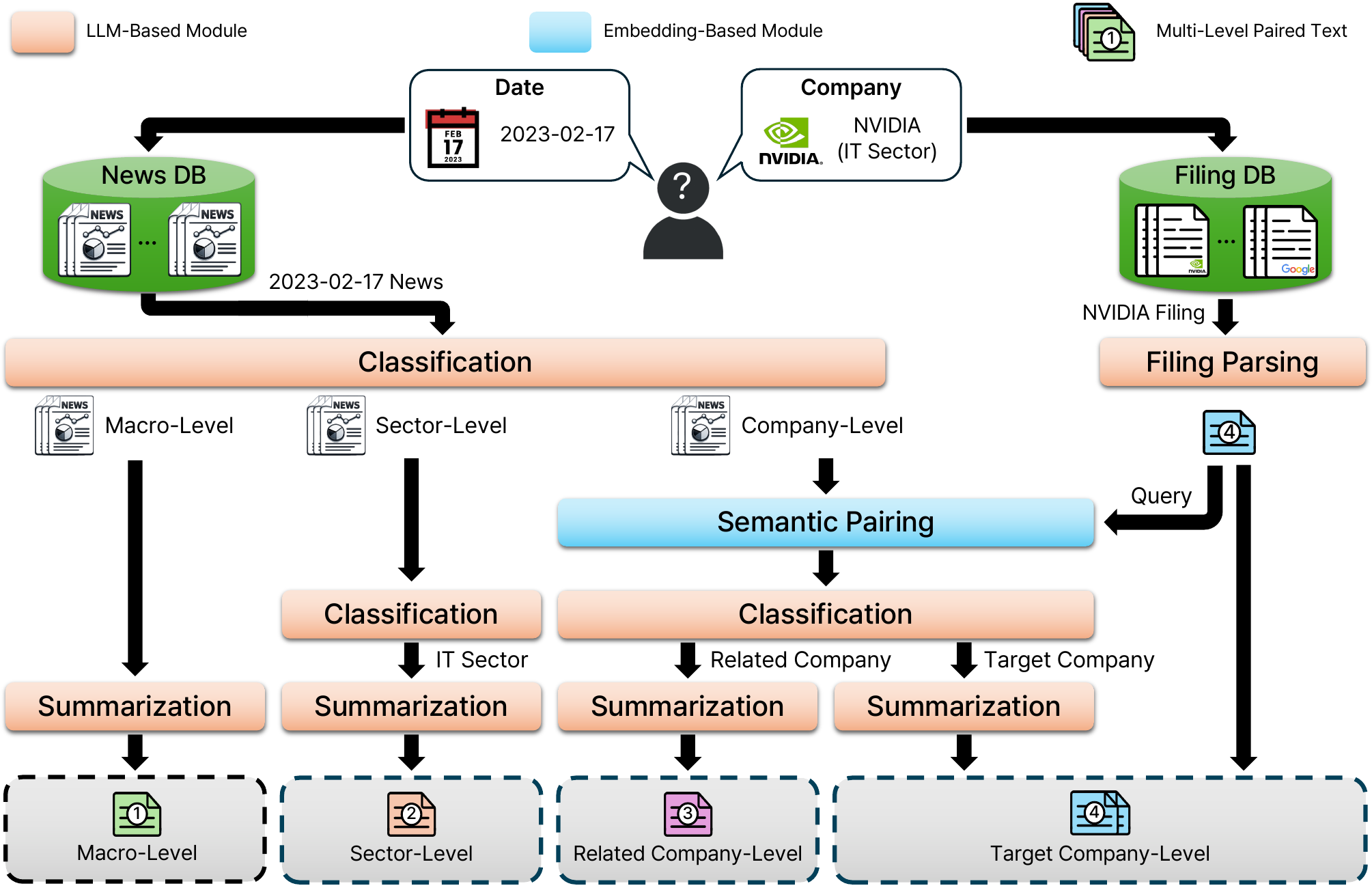}
    \vspace{-1.6em}
    \caption{
    Overview of the proposed semantic-based and multi-level pairing framework. Given a target company and a date, the framework retrieves semantically relevant texts at four levels (from \Circled{1} to \Circled{4}) and pairs them with the target company’s stock price on that date. \Circled{1} represents the paired text at the macro level, while \Circled{2} corresponds to the sector-level paired text. \Circled{3} and \Circled{4} denote the paired texts at the company level, where \Circled{3} captures the information of the related company and \Circled{4} represents the target company-specific information.
    }
    \label{fig:main}
\end{figure*}

The importance of time-series problems in finance, together with the growing interest in multimodal learning within the time-series community, has led to increased research efforts on constructing multimodal financial datasets as well as developing multimodal learning methods. However, the inherently complex nature of the financial domain makes the construction of text-paired time-series datasets particularly challenging. Specifically, in large-scale financial markets, individual components are interconnected to varying degrees, and these components often form clusters whose collective dynamics influence individual behaviors. For example, a company’s stock price is affected not only by events within the company itself, but also by developments involving related companies such as partners or competitors, as well as by broader factors including sector-wide trends and national macroeconomic conditions.

Several prior studies have attempted to construct text-paired time-series datasets specifically for stock price data in the financial domain~\cite{xu2018stock, dong2024fnspid}.
However, most existing approaches rely on simple keyword-based matching to pair individual companies’ stock prices with related texts, which leads to several limitations. First, such methods fail to capture texts that are semantically relevant to a company when the company is not explicitly mentioned. For instance, news articles discussing the construction of GPU data centers may be highly relevant to NVIDIA even if the company’s name does not appear in the text. Similarly, events reported in news articles about competitors or key business partners can materially affect a company despite the absence of a direct mention. Second, these simple pairing methods do not reflect the multi-level interactions inherent in financial markets. In reality, stock prices are influenced by an interplay of dynamics at multiple levels such as macro-level (country-level) and sector-level developments together with company-level factors. Keyword-based approaches are unable to adequately capture these intertwined, multi-level effects.


To overcome these limitations, we propose a 
\textit{semantic-based} and \textit{multi-level} pairing framework
and construct a new dataset, 
\textbf{FinTexTS} (\textbf{Fin}ancial \textbf{Tex}t-paired \textbf{T}ime-\textbf{S}eries Dataset),
applying the proposed approach to publicly available news from~\cite{dong2024fnspid}. \textbf{FinTexTS} is a large-scale text-paired stock price dataset 
covering 100 companies over a five-year period. As illustrated in Figure~\ref{fig:main}, we extract company-specific contextual information from SEC filings using an LLM-based parser and use this information to semantically align news articles with the corresponding companies. Our framework enables semantic pairing through embedding-based matching, leveraging an embedding model fine-tuned for the financial market domain. For multi-level pairing, we further employ an LLM-based classifier to assign each news article to its corresponding level.




In the experimental section, we conduct stock price forecasting as a pilot study to demonstrate the effectiveness of our well-paired text-TS dataset in the financial domain. Compared to forecasting results obtained using keyword-based paired text, our semantically paired text data provide more informative signals and lead to superior performance in our experiments. Furthermore, the forecasting performance improves when multi-level relationships are incorporated. In additional analyses, we show that applying our semantic-based and multi-level pairing framework to proprietary yet well-curated news sources leads to further improvements in forecasting performance compared to using publicly-available news. Specifically, we conduct experiments using the Machine Readable News (MRN) dataset provided by the London Stock Exchange Group (LSEG). By capturing more realistic and complex relationships in financial markets through our semantic-based and multi-level pairing framework and the \textbf{FinTexTS} dataset, we enable text–TS multimodal models to be trained and evaluated in more realistic settings, thereby contributing to future research in this area.





The main contributions are summarized as follows:
\setlist[itemize]{leftmargin=0.3cm, itemsep=5pt, topsep=0pt, partopsep=0pt}
\begin{itemize}

    \item In constructing financial text–TS multimodal datasets, we identify key limitations of existing keyword-based pairing methods, which fail to capture semantically relevant news without explicit company mentions and ignore multi-level market dynamics.

    
    \item To overcome these limitations, we propose a semantic-based and multi-level pairing framework. This framework extracts company-specific context from SEC filings, retrieves semantically relevant news via a fine-tuned embedding model, and classifies articles into multiple information levels using LLMs.
    



    \item By applying our framework to publicly-available news from~\cite{dong2024fnspid}, we construct \textbf{FinTexTS}, a large-scale text-paired stock price dataset. \textbf{FinTexTS} covers 100 publicly traded companies over a five-year period and includes paired textual data extracted from approximately 1 million news articles. 

    \item Through pilot stock price forecasting experiments, we demonstrate that our semantic-based and multi-level pairing strategy consistently outperforms keyword-based baselines. Furthermore, we show that applying our framework to proprietary but well-curated news sources yields additional performance improvements, highlighting the importance of both advanced pairing strategies and high-quality textual data.

    \item Our dataset is publicly available at \url{https://huggingface.co/datasets/EXAONE-BI/FinTexTS}. The fine-tuned embedding model is available at \url{https://huggingface.co/EXAONE-BI/FinTexTS-Embedding}. In addition, the code for our semantic-based and multi-level pairing framework, along with the pilot study implementation, is available at \url{https://github.com/leejaehoon2016/FinTexTS}.



    
\end{itemize}

\section{Related Works}
\subsection{Time-Series Forecasting in Finance}
Time-series forecasting is a fundamental task in machine learning, where neural networks aim to predict future observations by learning temporal dependencies from historical data. Since many financial signals, such as stock prices, exchange rates, and stock market indices (e.g., NASDAQ),
are inherently time-dependent, accurate forecasting of these signals is of significant importance in the financial domain~\cite{GIANTSIDI2025104719}. Among various financial forecasting problems, stock forecasting has emerged as one of the most extensively studied tasks in deep learning research. 
For example, DPA-STIFormer~\cite{yan2024doublepathadaptivecorrelationspatialtemporalinverted} improves stock forecasting performance by capturing implicit and dynamic inter-stock relationships through adaptively learned feature-driven correlations.

In addition, DoubleAdapt~\cite{zhao2024doubleadaptmetalearningapproachincremental} addresses the distribution shift problem in stock forecasting by explicitly modeling market non-stationarity and jointly adapting streaming stock data and model parameters under concept drift. Beyond direct price forecasting, prior work has also explored representation learning for capturing similarities among stocks, which can serve as a useful building block for downstream financial prediction tasks~\cite{hwang2023simstock}. Furthermore, recent work has also explored other financial prediction tasks; for instance, volatility surfaces combined with Vision Transformers have been used to predict realized volatility~\cite{soroka2026dataefficient}.

\begin{figure}[t]
\centering
\includegraphics[width=1.00\columnwidth]{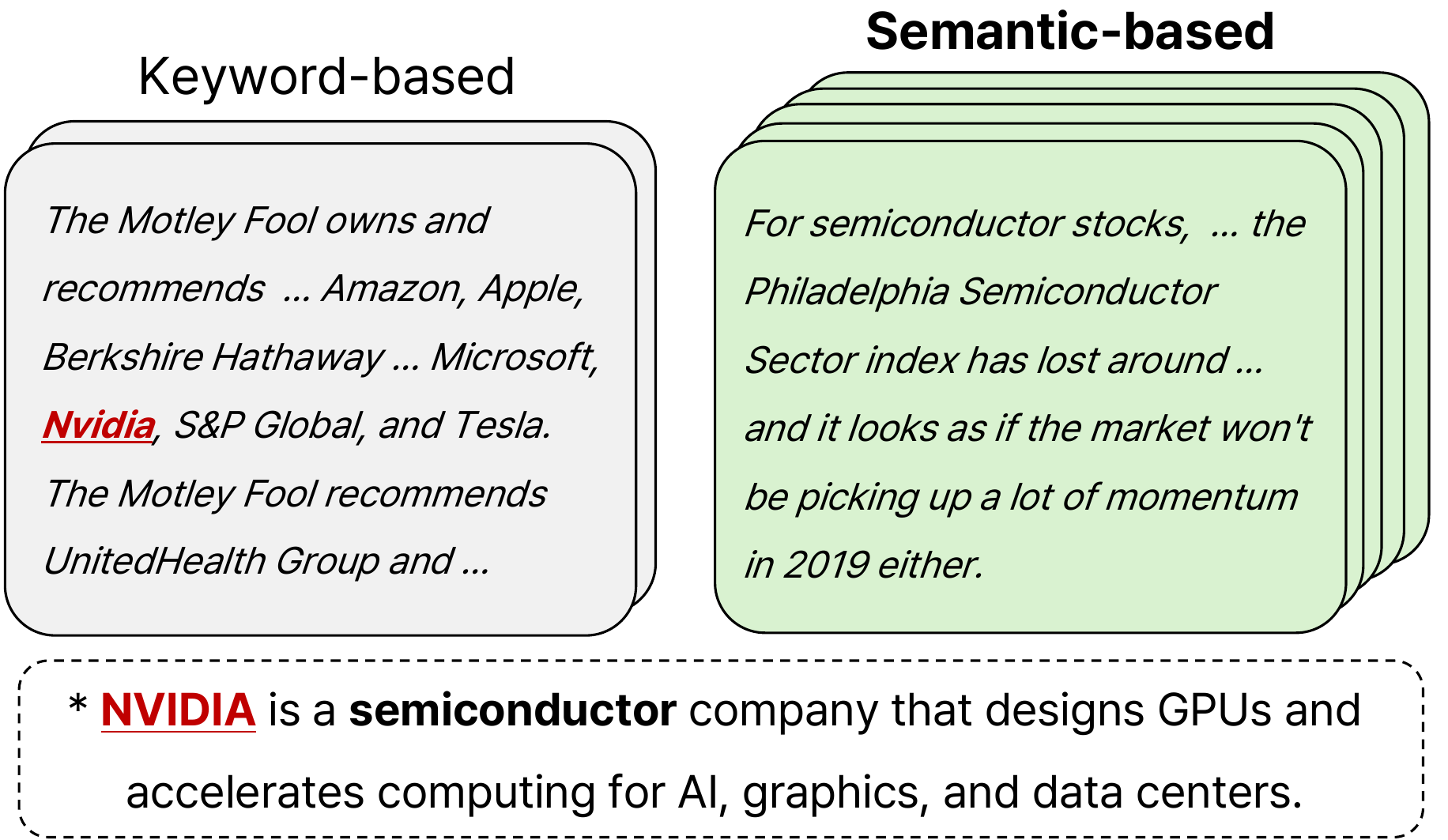} 
\vspace{-2em}
\caption{\textbf{Keyword-based vs. Semantic-based pairing method.}
Existing approaches rely on \textit{keyword-based} matching, whereas the proposed \emph{semantic-based} method enables the retrieval of more semantically relevant text even without explicit keywords.
}
\label{fig:keyword_vs_semantic}
\end{figure}
\subsection{Multimodal Time-Series Dataset}
Since the advent of multimodal models like vision-language models, the development of text-TS multimodal methods in time-series analysis has attracted many researchers' attention. This is because multimodal approaches have potential in that they can leverage additional information in addition to numerical data. However, constructing text-paired time-series datasets poses the following challenge: how to effectively align time-series data with relevant textual information. TimeMMD~\cite{liu2024timemmd} addresses this issue by using predefined text sets and keyword-based web searches. Nevertheless, this heuristic approach limits the diversity and quality of the paired text data, as texts retrieved from the same predefined sources tend to exhibit repetitive patterns. In the financial domain, FNSPID~\cite{dong2024fnspid} similarly pairs stock prices with news articles using a keyword-based matching strategy, linking news to stock prices when the target company is explicitly mentioned in the article. Similarly, prior work pairs stock price data with stock-specific tweets using a keyword-based matching strategy~\cite{xu2018stock}. However, as illustrated in Figure~\ref{fig:keyword_vs_semantic}, keyword-based selection can result in the inclusion of articles that are only weakly related-or even unrelated-to the target company. Moreover, it may fail to retrieve articles that are semantically relevant but do not explicitly mention the target company. These limitations lead to noisy and incomplete text-TS alignment. To address this issue, we propose a semantic-based and multi-level pairing framework, enabling more meaningful text-TS alignment.

\begin{figure*}[t]
    \centering
    \includegraphics[width=\linewidth]{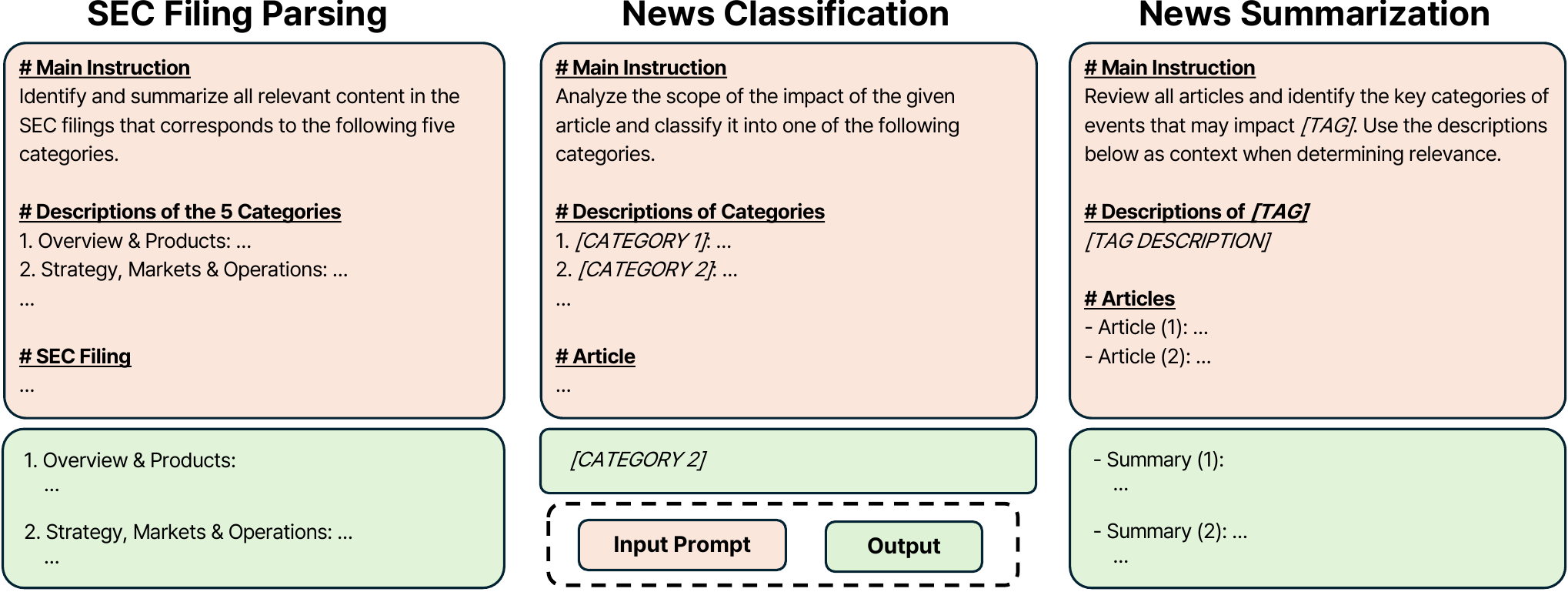}
    \vspace{-1.8em}
    \caption{LLM prompts used in our framework. Full version of LLM prompts are available in Appendix~\ref{appen:llm_prompt}.}
    \label{fig:prompt_samples}
\end{figure*}


\subsection{Financial Knowledge in LLMs}
Recent studies have demonstrated that LLMs possess substantial financial knowledge and reasoning capabilities. GPT-4 has been shown to analyze anonymized financial statements and predict future earnings direction with 60.35\% accuracy, outperforming human analysts (52.7\%) through genuine financial reasoning rather than memorization~\cite{kim2024financial}. 
Similarly, LLMs can accurately predict stock price movements from news headlines alone, suggesting that financial reasoning emerges as a capability in complex language models~\cite{lopezlira2023chatgpt}. 
FinBen~\cite{xie2024finben}, a comprehensive benchmark for evaluating LLMs on financial applications, systematically validated strong LLM performance across diverse financial tasks including sentiment analysis and stock trading. Furthermore, GPT-4 has demonstrated passing-level performance on Chartered Financial Analyst (CFA) examinations~\cite{callanan2024cfa}.
These findings suggest that LLMs possess sufficient financial domain knowledge to perform key tasks in our framework, including SEC filing parsing, news classification, and news summarization, thereby justifying our use of LLM-based modules.

\section{Method}

Our goal is to construct financial text–TS multimodal datasets using stock price data, which are widely used in financial time-series analysis. As paired textual data for stock prices, we leverage news articles that report facts and events potentially affecting a company, along with SEC filings that provide company-specific context about its current condition, such as financial statements and business overviews. Given a target company and a date, our framework pairs semantically relevant texts at four levels-macro, sector, related company, and target company-with the target company’s stock price on that date. There are four main components in our framework: an LLM-based SEC filing parser, LLM-based news classification, embedding-based news pairing, and LLM-based news summarization. Each component is described in detail from Section~\ref{sec:method:filing_parser} to Section~\ref{sec:method:summary}. We refer readers to Figure~\ref{fig:main} for an overview of the overall framework, Figure~\ref{fig:prompt_samples} for sample prompts used in our framework, and Appendix~\ref{appen:llm_prompt} for the full prompt set.



\subsection{LLM-Based SEC Filing Parsing}\label{sec:method:filing_parser}
An SEC filing is an official document that a company is legally required to submit to the U.S. Securities and Exchange Commission, disclosing its financial condition, operations, and significant events. Because SEC filings provide authoritative and company-specific information about a company’s current state, we use them as target company-level textual information paired with stock price data. However, SEC filing documents vary substantially in structure and format, making it necessary to employ a parser to extract standardized information.

To this end, we first define five categories that are commonly included in SEC filings and are particularly informative for describing a company’s current condition: \textit{i}) Overview \& Product, \textit{ii}) Strategy, Markets \& Operations, \textit{iii}) Governance \& Risks, \textit{iv}) Financial Information, and \textit{v}) Recent Events \& Catalysts, 
with detailed descriptions provided in Figure~\ref{fig:prompt_filing_parser}. We then design prompts that instruct an LLM to extract content corresponding to each category only when the relevant information is present in the filing, as shown in the first example in Figure~\ref{fig:prompt_samples}. Since SEC filings are not released on a daily basis, we apply forward-filling to each category to construct daily paired text–TS data. To maximize coverage and incorporate as much relevant information as possible, we leverage a diverse set of SEC filing types, including: S-1, S-4, F-1, F-4, S-1/A, F-1/A, F-4/A, 10-K, 20-F, 40-F, 10-K/A, 20-F/A, 40-F/A, 10-Q, 10-Q/A, 8-K, 6-K, and 8-K/A.

\subsection{LLM-Based News Classification}\label{sec:method:news_classification}
For multi-level pairing, we first classify news articles into three levels: macro-level, sector-level, and company-level. Macro-level news includes events that influence macroeconomic conditions at the national or global scale, such as fiscal policies, institutional actions by central financial authorities (e.g., the Federal Reserve), and developments in international relations. Sector-level news refers to events that do not affect the overall national economy but impact an entire industry sector\footnote{We adopt the Global Industry Classification Standard (GICS) sector in our framework, which is a well-known standardized industry classification system. The GICS framework consists of 11 sectors: Energy, Materials, Industrials, Consumer Discretionary, Consumer Staples, Health Care, Financials, Information Technology, Communication Services, Utilities, and Real Estate. Refer to Appendix~\ref{appen:llm_prompt} for the meaning of sectors.} or multiple companies within a specific sector. Finally, company-level news consists of articles that primarily affect a specific company or a small number of individual companies, rather than the broader economy or an entire sector. As illustrated in the second example of Figure~\ref{fig:prompt_samples}, we design prompts for LLM-based classification by providing a set of categories (e.g., macro, sector, and company) along with clear criteria specifying when an article should be assigned to each category.

In addition, we further refine sector-level news by assigning each article to a specific sector category. Specifically, we define a set of predefined sector categories based on the chosen industry classification standard---in our case, the Global Industry Classification Standard (GICS). An LLM-based classifier, similar to the one used above, is then employed to assign each news article to the most relevant sector category. After completing these steps, each company is paired, for a given date, with all macro-level news and the sector-level news corresponding to its assigned sector. For company-level news pairing, additional processing steps are performed as described in Section~\ref{sec:method:embedding}.








\subsection{Embedding-Based News Pairing}\label{sec:method:embedding}

After identifying company-level news articles, we aim to retrieve news articles that are semantically related to each target company. To enable semantic-based news pairing, we use the parsed contents of SEC filings as company-specific context and retrieve relevant news articles based on this information. 
Although a similar process to that in Section~\ref{sec:method:news_classification} could be implemented using LLMs by prompting the model with a list of companies and their corresponding contextual information and asking it to classify news accordingly, this approach does not scale well with the number of companies.
For instance, there are approximately 5,000 publicly listed companies in the U.S., and including contextual information for all of them results in an excessively large token count ($\approx$ 8M tokens).

To address this limitation, we propose an embedding-based news pairing method for identifying company-related news articles. We begin by adopting an existing pretrained embedding model and fine-tune it to better suit our task. Specifically, we hypothesize that effective retrieval of company-related news articles requires the ability to capture industry-level semantics, such as sectoral similarity in financial contexts. 

Accordingly, we leverage a subset of the news articles classified in Section~\ref{sec:method:news_classification} to fine-tune Linq-Embed-Mistral using its original architecture and hyperparameters. Specifically, we construct training pairs based on sector labels obtained from our LLM-based news classification pipeline, where news articles belonging to the same sector are treated as positive pairs and those from different sectors are treated as negative pairs. We then perform contrastive learning with TripletLoss such that semantically related news articles within the same sector have higher similarity scores, while articles from different sectors are pushed farther apart in the embedding space. The model is trained for 10 epochs with a batch size of 64 on 50,000 samples, consisting of 40,000 training samples and 10,000 validation samples.
Using the resulting embedding model, we perform semantic-based retrieval by treating each parsed filing component as a query and retrieving the top-$N$ company-level news articles published on the corresponding date.


After retrieval, we further apply a similar LLM-based classification approach to the one described in Section~\ref{sec:method:news_classification} to categorize the retrieved articles into three groups: target company-level, related company-level, and irrelevant. Target company-level news contains company-specific information that directly influences the target company, whereas related company-level news affects the target company indirectly through suppliers, partners, or competitors, without explicitly centering on the target company. To filter out noisy articles that may be included in the embedding-based retrieval results, we discard articles classified as irrelevant. After completing all steps in this section, the target company is paired with target company-level and related company-level news articles.

\subsection{LLM-Based News Summarization}\label{sec:method:summary}
After LLM-based news classification and embedding-based news pairing, news articles at each level are paired with the corresponding company. However, directly using raw news articles as paired textual inputs presents several challenges. 
First, a single news article often contains a mixture of relevant and irrelevant information, including peripheral details as well as the core events of interest.
Second, within each level, the paired news set may include both highly influential and less relevant articles. Third, redundant information may arise due to repeated coverage of similar events across multiple articles. These challenges motivate the need for a more compact and informative representation of the paired news articles.

To obtain such representations, we apply an LLM-based news summarization framework. As illustrated in the third example of Figure~\ref{fig:prompt_samples}, for each tag (e.g., U.S. macroeconomic for macro-level, one of the sector names for sector-level, or a company name for company-level), we provide the LLM with the corresponding context. Specifically, for the context, we use tag definitions for macro- and sector-level summarization, and the parsed contents of SEC filings for company-level summarization. We then ask the LLM to identify the top $N$ event categories that are expected to have the most significant impact on the company given this context and summarize the relevant news articles for each selected category.

\begin{figure*}[t]
    \centering
    \includegraphics[width=0.99\linewidth]{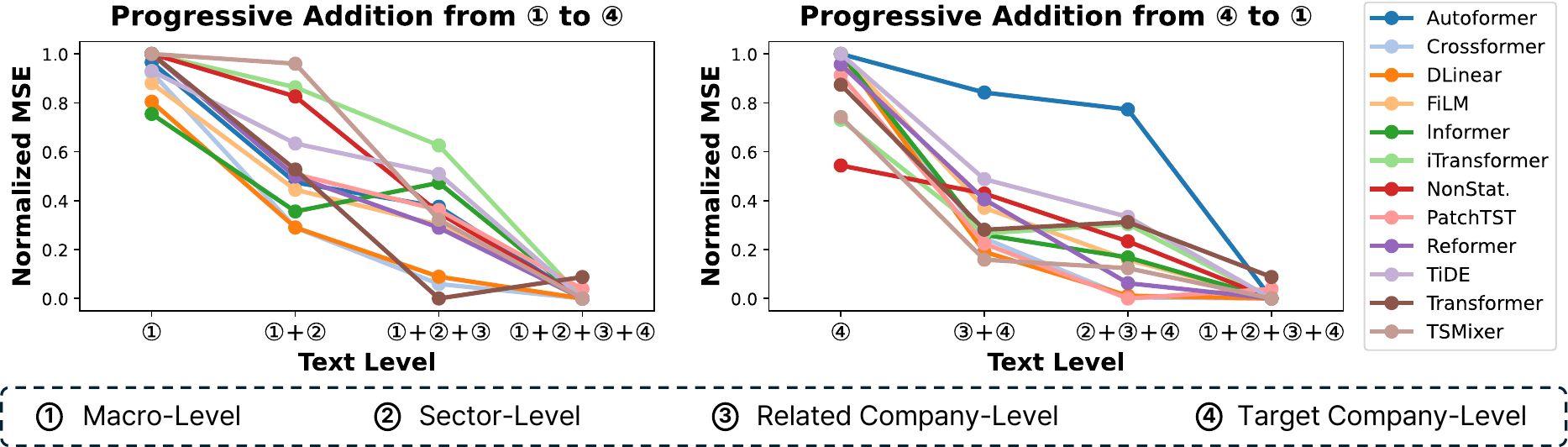}
    \vspace{-0.5em}
    \caption{Effect of Multi-Level Text Pairing on Forecasting Performance. In the left figure, textual information is progressively added from the macro level to the target company level to examine changes in forecasting performance, while the right figure shows performance changes when textual information is added in the reverse order.}
    \label{fig:multi-level}
\end{figure*}

\section{Experiment}
Using our proposed framework, we construct a dataset, \textbf{FinTexTS}, covering the top 100 companies by market capitalization over a five-year period from 2019 to 2023.
With \textbf{FinTexTS}, we present comprehensive experimental results to assess the effectiveness of our proposed semantic-based and multi-level pairing methods. First, Section~\ref{sec:exp:setup} describes the experimental setup. Then, Section~\ref{sec:exp:semantic-based} presents the experimental results for the semantic-based approach, while Section~\ref{sec:exp:multi-level} reports the results for the multi-level approach.

\subsection{Experimental Setup}\label{sec:exp:setup}
We construct a pilot study setting based on the experimental framework used in 
TimeMMD~\cite{liu2024timemmd}. The pilot study focuses on a text-augmented time-series forecasting task. In our experimental model, we first summarize the news content for each category using SBERT~\cite{reimers2019sentencebertsentenceembeddingsusing}, and then apply mean pooling across all categories and levels to obtain a final text representation. This text vector is passed through a projection layer and fused with the forecasting output from the time-series model to produce the final prediction.

To demonstrate the general applicability of our framework across different architectures, we evaluate it using 12 representative time-series forecasting models: Autoformer~\cite{wu2021autoformer}, Crossformer~\cite{zhang2023crossformer}, DLinear~\cite{zeng2023dlinear}, FiLM~\cite{zhou2022film}, Informer~\cite{zhou2021informer}, iTransformer~\cite{liu2024itransformer}, NonStationary Transformer~\cite{liu2022nonstationary}, PatchTST~\cite{nie2023patchtst}, Reformer~\cite{kitaev2020reformer}, TiDE~\cite{das2023tide}, Transformer~\cite{vaswani2017attention}, and TSMixer~\cite{chen2023tsmixer}. Following common practice in stock forecasting literature~\cite{dong2024fnspid}, we set the input horizon to 64 and the forecasting horizon to 3. For the price data, we use the open, high, low, and close values. The dataset is split into training, validation, and test sets spanning 3 years, 1 year, and 1 year, respectively.

For the LLM-based modules, we use gpt-4o-mini or gpt-5-mini via the OpenAI API~\cite{openai_api_reference}. For embedding-based news pairing, we fine-tune Linq-Embed-Mistral~\cite{choi2024linqembedmistraltechnicalreport} and retrieve 10 news articles per SEC filing component. In the LLM-based news summarization stage, the number of summarized categories is set to 5 for both macro-level and sector-level news, and 3 for company-level news, reflecting the relative volume of news at each level. Mean squared error (MSE) is used as the loss function during training and validation, while model performance is evaluated using MSE and mean absolute error (MAE). Since stock prices vary in scale across companies, we apply z-score normalization and compute MSE and MAE on the normalized values. All experiments are conducted on 100 companies with three random seeds, and we report the average results.

\begin{table}[]
    \centering
    \setlength{\tabcolsep}{5pt}  
    \renewcommand{\arraystretch}{1.0}
    \caption{Performance comparison of text-augmented stock forecasting under different text pairing strategies, including no textual information (w/o Text), keyword-based pairing (Keyword), and semantic-based pairing (Semantic). Note that `NonStat.' denotes the Non-stationary Transformer~\cite{liu2022nonstationary}.}
    \label{tab:semantic}
    \vspace{-0.8em}
    \resizebox{\linewidth}{!}{
    \begin{NiceTabular}{l|cc|cc|cc}
    \toprule
    \multirow{4}{*}{\quad \quad Model} & \multicolumn{2}{c|}{\multirow{2.5}{*}{w/o Text}} & \multicolumn{4}{c}{w/ Text} \\
    \cmidrule(lr){4-7}
     &  &  & \multicolumn{2}{c}{Keyword} & \multicolumn{2}{c}{Semantic} \\ 
     \cmidrule(lr){2-3} \cmidrule(lr){4-5} \cmidrule(lr){6-7}
     & MSE & MAE & MSE & MAE & MSE & MAE \\
    \midrule
    Autoformer   & 0.145 & 0.299 & 0.156 & 0.303 & \textbf{0.083} & \textbf{0.225} \\
    Crossformer  & 0.115 & 0.233 & 0.081 & 0.198 & \textbf{0.056} & \textbf{0.168} \\
    DLinear      & 0.025 & 0.120 & 0.035 & 0.141 & \textbf{0.022} & \textbf{0.110} \\
    FiLM         & 0.021 & 0.106 & 0.073 & 0.192 & \textbf{0.016} & \textbf{0.091} \\
    Informer     & 0.057 & 0.179 & 0.080 & 0.212 & \textbf{0.038} & \textbf{0.147} \\
    iTransformer & 0.027 & 0.124 & 0.097 & 0.222 & \textbf{0.021} & \textbf{0.109} \\
    NonStat.     & 0.043 & 0.155 & 0.070 & 0.193 & \textbf{0.028} & \textbf{0.127} \\
    PatchTST     & 0.018 & 0.097 & 0.089 & 0.209 & \textbf{0.017} & \textbf{0.096} \\
    Reformer     & 0.030 & 0.131 & 0.096 & 0.223 & \textbf{0.023} & \textbf{0.112} \\
    TiDE         & 0.028 & 0.127 & 0.081 & 0.207 & \textbf{0.021} & \textbf{0.107} \\
    Transformer  & 0.334 & 0.424 & 0.251 & 0.364 & \textbf{0.178} & \textbf{0.312} \\
    TSMixer      & 0.265 & 0.305 & 0.203 & 0.268 & \textbf{0.145} & \textbf{0.233} \\
    \bottomrule
    \end{NiceTabular}
    }
\vspace{-1.2em}
\end{table}

\subsection{Effect of Semantic-Based Pairing}\label{sec:exp:semantic-based}
In Table~\ref{tab:semantic}, we evaluate the effectiveness of semantic-based pairing, which constitutes one of the main components of our framework. As a baseline, we compare our approach against the keyword-based pairing method used in FNSPID~\cite{dong2024fnspid}. When evaluated on the top 100 companies by market capitalization, our semantic-based pairing consistently outperforms the keyword-based approach across all 12 forecasting models under both evaluation metrics. We attribute the weaker performance of the keyword-based approach to the following reasons: As illustrated in Figure~\ref{fig:keyword_vs_semantic}, keyword-based pairing can introduce news articles that are not truly relevant to the target company. Moreover, there are many days on which no news is paired with the target company, leading to inconsistencies in the paired textual inputs.\footnote{To handle this inconsistency, we apply forward filling.} Such irrelevant news and inconsistent pairing can significantly degrade forecasting performance. In addition, we compare our method with another baseline, where no textual information is incorporated. Our approach achieves superior performance in this setting as well, indicating that the paired textual information extracted by our framework provides meaningful signals that improve stock forecasting accuracy.

\begin{wraptable}{r}{0.22\textwidth}
\centering
\vspace{-3mm}
\caption{Backtesting results.}
\vspace{-1mm}
\label{tab:backtest}
\begin{tabular}{lcc}
\toprule
Method & Return & Sharpe \\
\midrule
Ours & \textbf{53.98\%} & \textbf{2.67} \\
FNSPID & 51.60\% & 2.57 \\
w/o Text & 51.50\% & 2.62 \\
\bottomrule
\end{tabular}
\vspace{-4mm}
\end{wraptable}
Beyond forecasting accuracy, we further evaluate the practical utility of the paired textual information through a backtesting experiment. Specifically, we construct an equal-weighted portfolio by selecting the top 20 stocks with the highest predicted 3-day returns and rebalancing every 3 days during the 2023 evaluation period. As shown in Table~\ref{tab:backtest}, our method achieves the best performance, obtaining a cumulative return of 53.98\% and a Sharpe ratio of 2.67, outperforming both the keyword-based FNSPID pairing method and the setting without textual information.

\subsection{Effect of Multi-Level Pairing}\label{sec:exp:multi-level}
Next, we evaluate the effectiveness of multi-level pairing, as shown in Figure~\ref{fig:multi-level}.
We progressively add paired textual information in the order of macro-level, sector-level, related company-level, and target company-level text, and observe the resulting changes in forecasting performance. We also perform the reverse procedure, adding paired texts in the opposite order, to examine the robustness of the observed trends.
To emphasize performance changes induced by the inclusion of textual information---rather than absolute performance differences across individual models---we report normalized MSE, obtained by applying min–max normalization to each model’s MSE across all seven evaluated text configurations. Accordingly, the y-axis in Figure~\ref{fig:multi-level} represents normalized MSE rather than raw MSE values. 
As shown in the figure, in both settings, model performance consistently improves as additional levels of textual information are incorporated. These results demonstrate that pairing textual information across multiple levels provides complementary signals that are beneficial for financial forecasting tasks such as stock price prediction.

\section{Analysis}
In this section, we present a range of analyses related to our framework. 
Sections~\ref{sec:analysis:retrieval_size} and~\ref{sec:analysis:fine-tune} conduct analyses of the embedding model used for semantic retrieval: Section~\ref{sec:analysis:retrieval_size} investigates the effect of the retrieval size $N$, while Section~\ref{sec:analysis:fine-tune} evaluates the improved retrieval performance achieved by our fine-tuning strategy.
Section~\ref{sec:analysis:case} provides a case study of text-paired forecasting.
Finally, Section~\ref{sec:analysis:close} examines performance changes when our framework is applied to high-quality closed-source news data, rather than publicly-available data, to construct paired text–TS datasets. Because our framework relies on LLMs, we conduct these analyses on a subset of 10 randomly selected companies from the 100 companies used in the main experiments, balancing the computational cost of LLMs with the generality of the results: AMD, BA, COST, DIS, GOOGL, INTC, NFLX, NVDA, T, and TSLA.

\begin{figure}
\centering
\includegraphics[width=1.00\columnwidth]{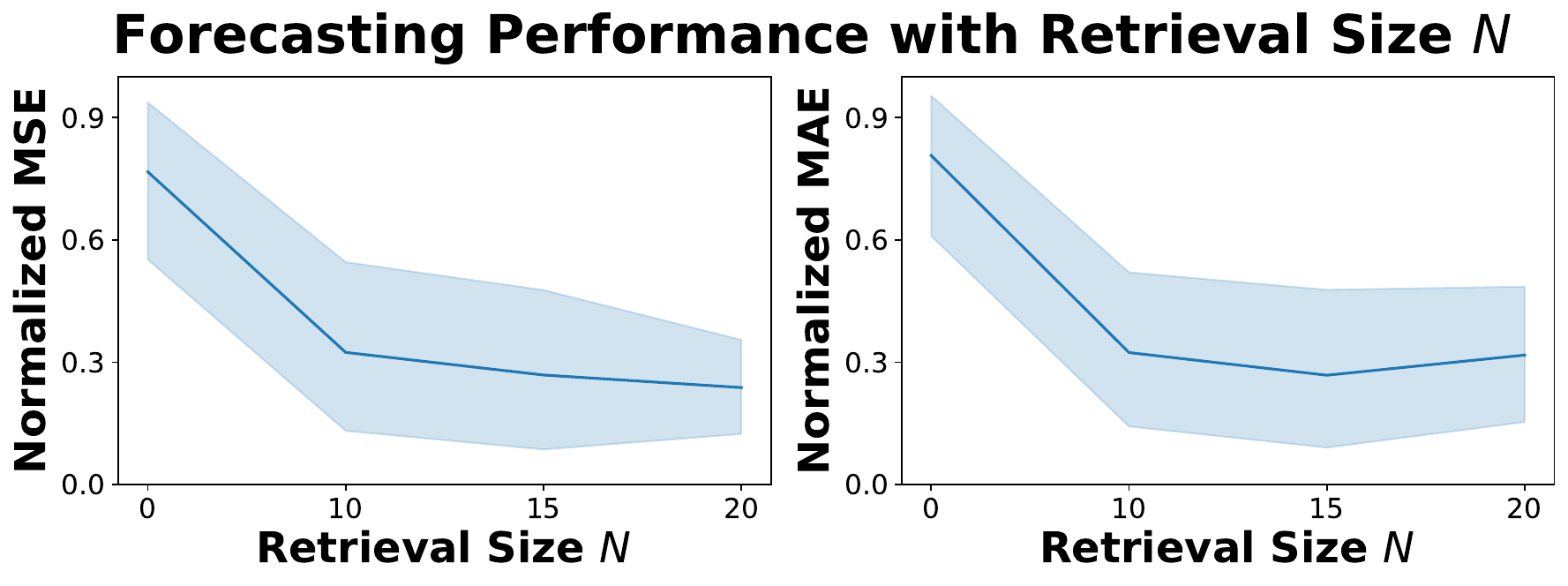} 
\vspace{-2em}
\caption{Sensitivity analysis of the retrieval size $N$ using the evaluation metrics MSE (left) and MAE (right).}
\label{fig:analysis:retrieval_size}
\vspace{-1em}
\end{figure}

\subsection{Sensitivity Analysis to Retrieval Size}\label{sec:analysis:retrieval_size}
As described in Section~\ref{sec:method:embedding}, we retrieve $N$ semantically related news articles for company-level pairing. In this section, we investigate the impact of the retrieval size $N$ through a sensitivity analysis. We set the search space of $N$ to $\{10, 15, 20\}$ and additionally include the case of $N=0$, where company-level news is not used. As shown in Figure~\ref{fig:multi-level}, to emphasize performance trends with respect to $N$ rather than absolute differences across individual forecasting models, we apply min–max normalization to the test MSE for each model across the four $N$ settings. We evaluate all 12 forecasting models and report the mean and standard deviation across models in the figure. As shown in Figure~\ref{fig:analysis:retrieval_size}, overall forecasting performance improves when $N>0$. Considering both the performance changes as $N$ increases and
the increased LLM usage cost
incurred by longer textual inputs, we choose $N=10$ as a balanced option between forecasting performance and computational cost.

\begin{table}[]
    \centering
    \setlength{\tabcolsep}{15pt}  
    \renewcommand{\arraystretch}{1.0}
    \caption{Retrieval performance comparison between using the embedding model as-is (Base) and our fine-tuned embedding model (Fine-Tuned). `IMP' denotes the improvement of our method over the baseline.}
    \label{tab:retrieve_performance}
    \vspace{-1em}
    \resizebox{\linewidth}{!}{
    \begin{NiceTabular}{l|cc|c}
    \toprule
     \multirow{2.5}{*}{Company} &
     \multicolumn{2}{c}{Embedding Model} & \multirow{2.5}{*}{IMP} \\
     \cmidrule(lr){2-3}
     & Base & Fine-Tuned &  \\
    \midrule
    AMD   & 62.3\% & \textbf{65.5\%} & +3.2\%p \\
    BA    & 67.9\% & \textbf{69.6\%} & +1.7\%p \\
    COST  & \textbf{66.8\%} & \textbf{66.8\%} & +0.0\%p \\
    DIS   & 90.7\% & \textbf{92.2\%} & +1.5\%p \\
    GOOGL & 86.0\% & \textbf{87.4\%} & +1.4\%p \\
    INTC  & 74.4\% & \textbf{76.8\%} & +2.4\%p \\
    NFLX  & 67.6\% & \textbf{73.5\%} & +5.9\%p \\
    NVDA  & 70.0\% & \textbf{74.2\%} & +4.2\%p \\
    T     & 78.5\% & \textbf{80.2\%} & +1.7\%p \\
    TSLA  & 74.7\% & \textbf{77.4\%} & +2.7\%p \\
    \bottomrule
    \end{NiceTabular}
    }
    
\end{table}

\subsection{Effect of Fine-Tuned Embedding Model}\label{sec:analysis:fine-tune}
As discussed in Section~\ref{sec:method:embedding}, we fine-tune an embedding model using the sector-level news classification information obtained in Section~\ref{sec:method:news_classification}. In this section, we investigate whether our fine-tuning strategy improves news retrieval performance. In Table~\ref{tab:retrieve_performance}, we compare the retrieval performance of the base embedding model with that of our fine-tuned model. As mentioned in Section~\ref{sec:exp:setup}, we use Linq-Embed-Mistral as the base embedding model in our experiments. To evaluate retrieval quality, we adopt the hit-rate metric, which measures the probability that at least one target company-level news article is included among the retrieved results. The results show consistent improvements across the evaluated metric for the 10 companies. These findings indicate that our embedding fine-tuning strategy enhances document retrieval performance and is beneficial for semantic-based news pairing.

%
%

\begin{figure}[t]
\centering
\includegraphics[width=1.00\columnwidth]{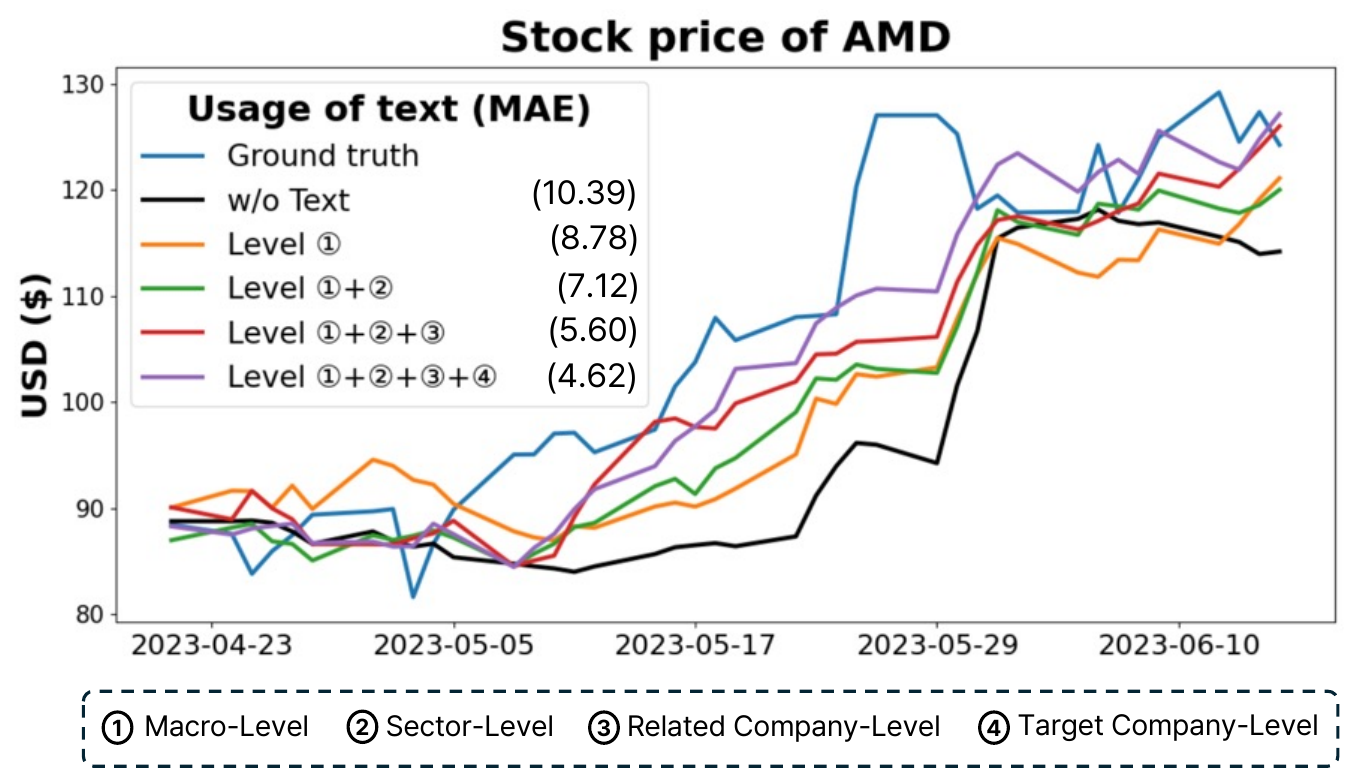} 
\vspace{-1.8em}
\caption{Effect of multi-level text on AMD (Advanced Micro Devices, Inc.) stock forecasting.}
\label{fig:analysis:case_study}
\end{figure}

\begin{figure*}[t]
    \centering
    \includegraphics[width=0.99\linewidth]{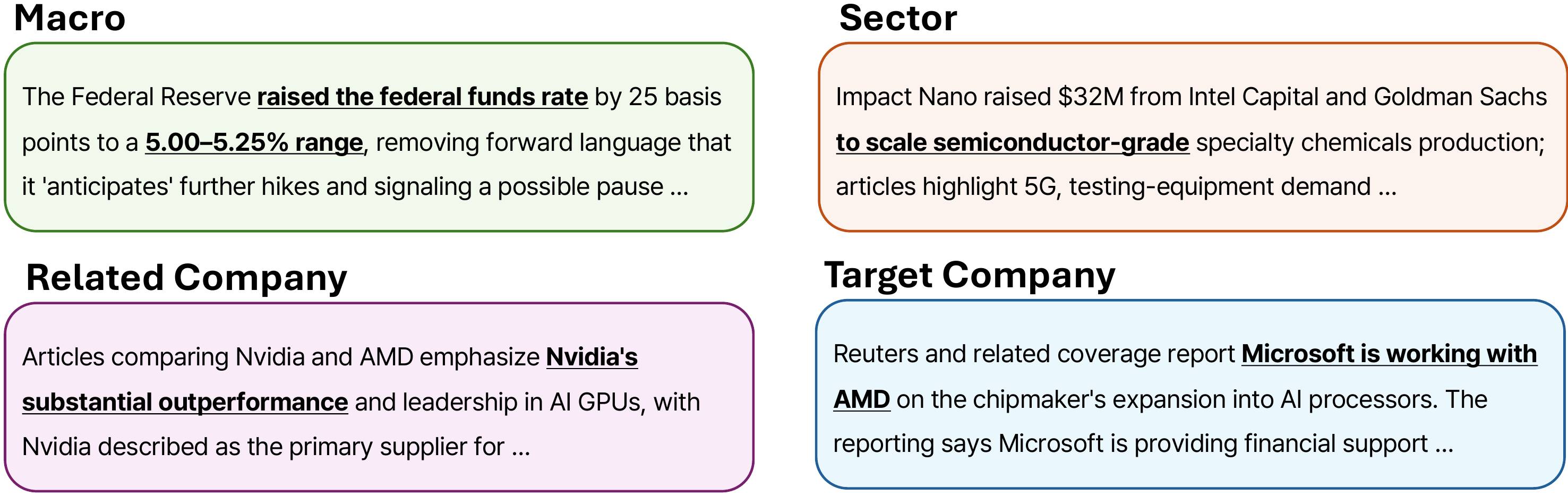}
    \vspace{-0.7em}
    \caption{Multi-level paired text examples for AMD, which is a U.S. semiconductor company that designs high-performance CPUs and GPUs for PCs, data centers, and AI applications.}
    \label{fig:analysis:case_text}
\end{figure*}

\subsection{Case Study}\label{sec:analysis:case}
This section presents a case study to examine whether the multi-level paired texts constructed by our framework contribute to improved forecasting performance. In Figure~\ref{fig:analysis:case_study}, we report the close-price forecasting results of Non-stationary Transformer for a major U.S. company, Advanced Micro Devices, Inc. (AMD). The figure shows two months of forecasts generated using a rolling-window strategy. As shown in the figure, forecasting performance gradually improves as additional textual information from multiple levels is incorporated. This trend is also reflected in the MAE scores, which consistently decrease as more text is added.

In Figure~\ref{fig:analysis:case_text}, we present an example of paired texts at each level for AMD on 2023-05-04. The paired texts capture diverse contextual information surrounding the target company. At the macro level, the paired news discusses the government’s tightening monetary stance, whereas the sector-level news highlights favorable tailwinds for the semiconductor industry. At the company level, related company-level news includes articles about competitors’ outperformance, while target company-level news contains reports on AMD’s collaboration with Microsoft. These examples show that our dataset reflects multi-level events and signals that may jointly influence a company’s stock price dynamics.

\begin{table}[]
    \centering
    \renewcommand{\arraystretch}{1.0}
    \setlength{\tabcolsep}{9pt}  
    \caption{Comparison of forecasting performance with publicly-available and proprietary news sources.}
    \vspace{-0.5em}
    \label{tab:close}
    \resizebox{\linewidth}{!}{
    \begin{NiceTabular}{l|cc|cc}
    \toprule
    \multirow{2.5}{*}{\quad \ \ Model} & \multicolumn{2}{c|}{Publicly-Available} & \multicolumn{2}{c}{Proprietary} \\
    \cmidrule(lr){2-3} \cmidrule(lr){4-5}
     & MSE & MAE & MSE & MAE \\
    \midrule
    Autoformer    & 0.084 & 0.231 & \textbf{0.070} & \textbf{0.204} \\
    Crossformer   & 0.038 & 0.143 & \textbf{0.029} & \textbf{0.123} \\
    DLinear       & 0.024 & 0.112 & \textbf{0.022} & \textbf{0.107} \\
    FiLM          & 0.018 & 0.093 & \textbf{0.017} & \textbf{0.090} \\
    Informer      & 0.041 & 0.150 & \textbf{0.034} & \textbf{0.136} \\
    iTransformer  & 0.023 & 0.112 & \textbf{0.022} & \textbf{0.108} \\
    NonStat.      & 0.032 & 0.133 & \textbf{0.029} & \textbf{0.125} \\
    PatchTST      & 0.019 & 0.097 & \textbf{0.018} & \textbf{0.094} \\
    Reformer      & 0.024 & 0.115 & \textbf{0.022} & \textbf{0.110} \\
    TiDE          & 0.026 & 0.116 & \textbf{0.025} & \textbf{0.115} \\
    Transformer   & 0.117 & 0.266 & \textbf{0.082} & \textbf{0.220} \\
    TSMixer       & 0.060 & 0.168 & \textbf{0.037} & \textbf{0.136} \\
    \bottomrule
    \end{NiceTabular}
    }
    \vspace{-1em}
\end{table}


\subsection{Proprietary \& Well-Curated News Source}\label{sec:analysis:close}
We use publicly-available news to construct our \textbf{FinTexTS} dataset. In this section, we investigate the effect of using a proprietary but well-curated news source. We hypothesize that proprietary news can provide higher-quality textual content and broader coverage than publicly-available news, which may lead to richer paired texts and improved forecasting performance. To evaluate this, we apply our framework to the Machine Readable News (MRN) dataset provided by the London Stock Exchange Group (LSEG) and construct paired text data accordingly. As shown in Table~\ref{tab:close}, we observe that using the proprietary data yields performance improvements over publicly-available news in most experimental settings. These results suggest that leveraging proprietary, well-curated news sources, such as LSEG MRN, can be beneficial for constructing higher-quality paired text data and further improving forecasting results.

\begin{table}[t]
    \centering
    \renewcommand{\arraystretch}{1.2}
    \setlength{\tabcolsep}{2pt}
    \caption{LLM-based evaluation of paired text quality, where each criterion is scored on a 1--5 scale (higher is better).}
    \vspace{-0.5em}
    \label{tab:llm_eval}
    \resizebox{\linewidth}{!}{
    \begin{NiceTabular}{cc|ccc|c}
    \toprule
    \multicolumn{2}{c|}{\multirow{2}{*}{Method}} 
    & \multicolumn{3}{c|}{Criteria} 
    & \multirow{2}{*}{Overall} \\
    \cmidrule(lr){3-5}
    \multicolumn{2}{c|}{} 
    & Coverage & Relatedness & Diversity & \\
    \midrule
    \multicolumn{2}{c|}{FNSPID} 
    & 1.17 & 1.70 & 1.18 & 1.35 \\
    \midrule
    \multirow{2}{*}{FinTexTS} 
    & Open-Source & 3.99 & 4.16 & 4.17 & 4.11 \\
    & Proprietary & \textbf{4.27} & \textbf{4.23} & \textbf{4.60} & \textbf{4.36} \\
    \bottomrule
    \end{NiceTabular}
    }
    \vspace{-1em}
\end{table}

\subsection{LLM Evaluation of Paired Text Quality}\label{sec:analysis:llm_eval}

To further evaluate the quality of the paired textual data constructed by our framework, we conduct an additional analysis using an LLM-as-a-judge evaluation framework. Specifically, we compare the paired texts generated by the keyword-based pairing strategy of FNSPID, our framework using publicly-available news sources, and our framework using the proprietary LSEG MRN news source. We design an evaluation rubric consisting of three criteria: \textit{i)} coverage, which measures how broadly the paired texts capture important factors affecting stock prices across macro, sector, and company levels; \textit{ii)} relatedness, which evaluates how directly the content is relevant to the target company; and \textit{iii)} diversity, which measures the variety of perspectives and topics covered by the paired texts. Each criterion is scored on a 1--5 scale, where higher scores indicate better quality.

As shown in Table~\ref{tab:llm_eval}, our semantic-based and multi-level pairing framework substantially improves paired text quality compared to the keyword-based FNSPID approach across all evaluation criteria. Furthermore, using the proprietary LSEG MRN news source leads to additional improvements in all metrics, supporting our hypothesis that well-curated proprietary news sources can provide richer contextual information and broader coverage for financial text--TS pairing.





\section{Conclusion}
In constructing financial text-paired time-series datasets, this paper has the following key contributions: \textit{i)} identifying the challenges caused by the complex nature in financial markets, \textit{ii)} proposing a semantic-based and multi-level pairing framework, \textit{iii)} constructing a large-scale financial text-time series (text-TS) dataset with the proposed method. In financial markets, individual entities are interconnected to varying degrees (e.g., through supplier, partner, or competitor relationships). These entities are also influenced by broader collective dynamics such as national macroeconomic conditions and sector-wide trends.

Motivated by the fact that such complex dependencies are not well captured by existing keyword-based text–TS pairing methods, we propose a text–TS pairing framework that incorporates two key components: semantic-based and multi-level pairing. First, we extract company-specific context from SEC filings. Based on this context, we perform embedding-based semantic pairing to retrieve news articles relevant to the target company even when the company is not explicitly mentioned. Furthermore, through multi-level pairing, we organize paired texts into four levels: macro-level, sector-level, related company-level, and target company-level. Using this framework, we build \textbf{FinTexTS}, a large-scale text-paired stock price dataset spanning 100 companies over five years, with paired texts derived from approximately one million news articles.

In our experiments, we demonstrate the effectiveness of \textbf{FinTexTS} 
with multimodal forecasting models for stock forecasting.
Results show that both semantic-based and multi-level pairing consistently improve forecasting performance across diverse model architectures. In addition, we provide extensive analyses, including sensitivity analyses on retrieval size, evaluations of our embedding fine-tuning strategy, qualitative case studies, and experiments using proprietary news sources. For future work, we leave the development of text–TS multimodal architectures and learning methods specifically designed to better exploit multi-level textual information as an important direction for further research.

\textbf{Ethics Statement.} This work uses news data sourced from the FNSPID dataset~\cite{dong2024fnspid}, which was collected in accordance with established ethical standards. The proposed dataset is intended solely for research purposes, and users are encouraged to consider ethical implications and potential misuse when applying it in real-world financial contexts.

\textbf{Limitations.} While \textbf{FinTexTS} is constructed at a large scale, our dataset is currently limited to a subset of U.S. publicly traded companies and a specific temporal range. Nevertheless, we believe the dataset remains reasonably representative due to two main factors: \textit{i)} the inclusion of major publicly traded companies that account for a substantial portion of overall market activity, and \textit{ii)} the large-scale news corpus consisting of approximately one million news articles. We acknowledge that these design choices may introduce potential geographic and temporal biases, and we plan to further expand the dataset coverage and investigate such limitations in future work.



\bibliographystyle{ACM-Reference-Format}
\bibliography{reference}

@inproceedings{nie2023patchtst,
  title={A Time Series is Worth 64 Words: Long-term Forecasting with Transformers},
  author={Nie, Yuqi and Nguyen, Nam H and Sinthong, Phanwadee and Kalagnanam, Jayant},
  booktitle={International Conference on Learning Representations},
  year={2023}
}

@inproceedings{liu2024itransformer,
  title={iTransformer: Inverted Transformers Are Effective for Time Series Forecasting},
  author={Liu, Yong and Hu, Tengge and Zhang, Haoran and Wu, Haixu and Wang, Shiyu and Ma, Lintao and Long, Mingsheng},
  booktitle={International Conference on Learning Representations},
  year={2024}
}

@inproceedings{xu2018stock,
  title={Stock Movement Prediction from Tweets and Historical Prices},
  author={Xu, Yumo and Cohen, Shay B},
  booktitle={Proceedings of the 56th Annual Meeting of the Association for Computational Linguistics (Volume 1: Long Papers)},
  pages={1970--1979},
  year={2018}
}

@inproceedings{radford2021clip,
  title={Learning Transferable Visual Models from Natural Language Supervision},
  author={Radford, Alec and Kim, Jong Wook and Hallacy, Chris and Ramesh, Aditya and Goh, Gabriel and Agarwal, Sandhini and Sastry, Girish and Askell, Amanda and Mishkin, Pamela and Clark, Jack and others},
  booktitle={International Conference on Machine Learning},
  pages={8748--8763},
  year={2021},
  organization={PMLR}
}

@inproceedings{li2022blip,
  title={BLIP: Bootstrapping Language-Image Pre-training for Unified Vision-Language Understanding and Generation},
  author={Li, Junnan and Li, Dongxu and Xiong, Caiming and Hoi, Steven},
  booktitle={International Conference on Machine Learning},
  pages={12888--12900},
  year={2022},
  organization={PMLR}
}

@inproceedings{liu2023llava,
  title={Visual Instruction Tuning},
  author={Liu, Haotian and Li, Chunyuan and Wu, Qingyang and Lee, Yong Jae},
  booktitle={Advances in Neural Information Processing Systems},
  volume={36},
  year={2023}
}

@inproceedings{liu2024timemmd,
  title={Time-MMD: Multi-Domain Multimodal Dataset for Time Series Analysis},
  author={Liu, Haoxin and Xu, Shangqing and Zhao, Zhiyuan and Kong, Lingkai and Kamarthi, Harshavardhan and Sasanur, Aditya B and Sharma, Megha and Cui, Jiaming and Wen, Qingsong and Zhang, Chao and Prakash, B Aditya},
  booktitle={Advances in Neural Information Processing Systems},
  volume={37},
  year={2024}
}

@inproceedings{dong2024fnspid,
  author    = {Dong, Zihan and Fan, Xinyu and Peng, Zhiyuan},
  title     = {{FNSPID}: A Comprehensive Financial News Dataset in Time Series},
  year      = {2024},
  booktitle = {Proceedings of the 30th ACM SIGKDD Conference on Knowledge Discovery and Data Mining},
  series    = {KDD '24},
  pages     = {4918--4927},
  publisher = {Association for Computing Machinery},
  address   = {New York, NY, USA},
  doi       = {10.1145/3637528.3671629}
}

@article{seo2025air,
  title={Adaptive Information Routing for Multimodal Time Series Forecasting},
  author={Seo, Jun and others},
  journal={arXiv preprint arXiv:2512.10229},
  year={2025}
}

@article{GIANTSIDI2025104719,
title = {Deep learning for financial forecasting: A review of recent trends},
journal = {International Review of Economics \& Finance},
volume = {104},
pages = {104719},
year = {2025},
issn = {1059-0560},
doi = {https://doi.org/10.1016/j.iref.2025.104719},
url = {https://www.sciencedirect.com/science/article/pii/S1059056025008822},
author = {Sofia Giantsidi and Claudia Tarantola},
}

@misc{yan2024doublepathadaptivecorrelationspatialtemporalinverted,
      title={Double-Path Adaptive-correlation Spatial-Temporal Inverted Transformer for Stock Time Series Forecasting}, 
      author={Wenbo Yan and Ying Tan},
      year={2024},
      eprint={2409.15662},
      archivePrefix={arXiv},
      primaryClass={cs.LG},
      url={https://arxiv.org/abs/2409.15662}, 
}

@misc{zhao2024doubleadaptmetalearningapproachincremental,
      title={DoubleAdapt: A Meta-learning Approach to Incremental Learning for Stock Trend Forecasting}, 
      author={Lifan Zhao and Shuming Kong and Yanyan Shen},
      year={2024},
      eprint={2306.09862},
      archivePrefix={arXiv},
      primaryClass={q-fin.ST},
      url={https://arxiv.org/abs/2306.09862}, 
}

@misc{kim2024financial,
      title={Financial Statement Analysis with Large Language Models}, 
      author={Alex Kim and Maximilian Muhn and Valeri Nikolaev},
      year={2024},
      eprint={2407.17866},
      archivePrefix={arXiv},
      primaryClass={q-fin.ST},
      url={https://arxiv.org/abs/2407.17866}, 
}

@misc{lopezlira2023chatgpt,
      title={Can ChatGPT Forecast Stock Price Movements? Return Predictability and Large Language Models}, 
      author={Alejandro Lopez-Lira and Yuehua Tang},
      year={2023},
      eprint={2304.07619},
      archivePrefix={arXiv},
      primaryClass={q-fin.ST},
      url={https://arxiv.org/abs/2304.07619}, 
}

@inproceedings{
soroka2026dataefficient,
title={Data-Efficient Realized Volatility Forecasting with Vision Transformers},
author={Emi Soroka and Artem Arzyn},
booktitle={NeurIPS 2025 Workshop: Generative AI in Finance},
year={2026},
url={https://openreview.net/forum?id=0bxRD79zv1}
}

@inproceedings{xie2024finben,
  title={FinBen: A Holistic Financial Benchmark for Large Language Models},
  author={Xie, Qianqian and Han, Weiguang and Chen, Zhengyu and Xiang, Ruoyu and Zhang, Xiao and He, Yueru and Xiao, Mengxi and Li, Dong and Dai, Yongfu and Feng, Duanyu and others},
  booktitle={Advances in Neural Information Processing Systems},
  volume={37},
  year={2024}
}

@inproceedings{callanan2024cfa,
  title={Can GPT models be Financial Analysts? An Evaluation of ChatGPT and GPT-4 on mock CFA Exams},
  author={Callanan, Ethan and Mbakwe, Amarachi and Papadimitriou, Antony and Pei, Yulong and Sibue, Mathieu and Zhu, Xiaodan and Ma, Zhiqiang and Liu, Xiaomo and Shah, Sameena},
  booktitle={FinNLP-AgentScen Workshop @ IJCAI},
  year={2024}
}

@inproceedings{wu2021autoformer,
  title={Autoformer: Decomposition Transformers with Auto-Correlation for Long-Term Series Forecasting},
  author={Wu, Haixu and Xu, Jiehui and Wang, Jianmin and Long, Mingsheng},
  booktitle={Advances in Neural Information Processing Systems},
  volume={34},
  pages={22419--22430},
  year={2021}
}

@inproceedings{zhang2023crossformer,
  title={Crossformer: Transformer Utilizing Cross-Dimension Dependency for Multivariate Time Series Forecasting},
  author={Zhang, Yunhao and Yan, Junchi},
  booktitle={International Conference on Learning Representations},
  year={2023}
}

@inproceedings{zeng2023dlinear,
  title={Are Transformers Effective for Time Series Forecasting?},
  author={Zeng, Ailing and Chen, Muxi and Zhang, Lei and Xu, Qiang},
  booktitle={Proceedings of the AAAI Conference on Artificial Intelligence},
  volume={37},
  number={9},
  pages={11121--11128},
  year={2023}
}

@inproceedings{zhou2022film,
  title={FiLM: Frequency improved Legendre Memory Model for Long-term Time Series Forecasting},
  author={Zhou, Tian and Ma, Ziqing and Wen, Qingsong and Sun, Liang and Yao, Tao and Yin, Wotao and Jin, Rong},
  booktitle={Advances in Neural Information Processing Systems},
  volume={35},
  pages={12677--12690},
  year={2022}
}

@inproceedings{zhou2021informer,
  title={Informer: Beyond Efficient Transformer for Long Sequence Time-Series Forecasting},
  author={Zhou, Haoyi and Zhang, Shanghang and Peng, Jieqi and Zhang, Shuai and Li, Jianxin and Xiong, Hui and Zhang, Wancai},
  booktitle={Proceedings of the AAAI Conference on Artificial Intelligence},
  volume={35},
  number={12},
  pages={11106--11115},
  year={2021}
}

@inproceedings{liu2022nonstationary,
  title={Non-stationary Transformers: Exploring the Stationarity in Time Series Forecasting},
  author={Liu, Yong and Wu, Haixu and Wang, Jianmin and Long, Mingsheng},
  booktitle={Advances in Neural Information Processing Systems},
  volume={35},
  pages={9881--9893},
  year={2022}
}

@inproceedings{kitaev2020reformer,
  title={Reformer: The Efficient Transformer},
  author={Kitaev, Nikita and Kaiser, {\L}ukasz and Levskaya, Anselm},
  booktitle={International Conference on Learning Representations},
  year={2020}
}

@article{das2023tide,
  title={Long-term Forecasting with TiDE: Time-series Dense Encoder},
  author={Das, Abhimanyu and Kong, Weihao and Leber, Andrew and Mathur, Rajat and Sen, Rajat and Yu, Rose},
  journal={Transactions on Machine Learning Research},
  year={2023}
}

@inproceedings{vaswani2017attention,
  title={Attention is All You Need},
  author={Vaswani, Ashish and Shazeer, Noam and Parmar, Niki and Uszkoreit, Jakob and Jones, Llion and Gomez, Aidan N and Kaiser, {\L}ukasz and Polosukhin, Illia},
  booktitle={Advances in Neural Information Processing Systems},
  volume={30},
  year={2017}
}

@article{chen2023tsmixer,
  title={TSMixer: An All-MLP Architecture for Time Series Forecasting},
  author={Chen, Si-An and Li, Chun-Liang and Yoder, Nate and Arik, Sercan O. and Pfister, Tomas},
  journal={Transactions on Machine Learning Research},
  year={2023}
}

@inproceedings{luo2018neural,
  title={A Neural Stochastic Volatility Model},
  author={Luo, Rui and Zhang, Weinan and Xu, Xiaojun and Wang, Jun},
  booktitle={Proceedings of the AAAI Conference on Artificial Intelligence},
  volume={32},
  number={1},
  pages={6401--6408},
  year={2018}
}

@inproceedings{nakagawa2019deep,
  title={Deep Recurrent Factor Model: Interpretable Non-Linear and Time-Varying Multi-Factor Model},
  author={Nakagawa, Kei and Ito, Tomoki and Abe, Masaya and Izumi, Kiyoshi},
  booktitle={AAAI-19 Workshop on Network Interpretability for Deep Learning},
  year={2019}
}

@inproceedings{zhu2025fincast,
  title={FinCast: A Foundation Model for Financial Time-Series Forecasting},
  author={Zhu, Zhuohang and Chen, Haodong and Qu, Qiang and Chung, Vera},
  booktitle={Proceedings of the 34th ACM International Conference on Information and Knowledge Management (CIKM)},
  year={2025}
}

@inproceedings{li2024master,
  title={MASTER: Market-Guided Stock Transformer for Stock Price Forecasting},
  author={Li, Tong and Liu, Zhaoyang and Shen, Yanyan and Wang, Xue and Chen, Haokun and Huang, Sen},
  booktitle={Proceedings of the AAAI Conference on Artificial Intelligence},
  volume={38},
  number={1},
  pages={162--170},
  year={2024}
}

@inproceedings{yoo2021accurate,
  title={Accurate Multivariate Stock Movement Prediction via Data-Axis Transformer with Multi-Level Contexts},
  author={Yoo, Jaemin and Soun, Yejun and Park, Yong-chan and Kang, U},
  booktitle={Proceedings of the 27th ACM SIGKDD Conference on Knowledge Discovery \& Data Mining},
  pages={2037--2045},
  year={2021}
}

@inproceedings{hwang2023simstock,
  title={SimStock: Representation Model for Stock Similarities},
  author={Hwang, Yoontae and Lee, Junhyeong and Kim, Daham and Noh, Seunghwan and Hong, Joohwan and Lee, Yongjae},
  booktitle={Proceedings of the Fourth ACM International Conference on AI in Finance (ICAIF '23)},
  year={2023}
}

@misc{choi2024linqembedmistraltechnicalreport,
      title={Linq-Embed-Mistral Technical Report}, 
      author={Chanyeol Choi and Junseong Kim and Seolhwa Lee and Jihoon Kwon and Sangmo Gu and Yejin Kim and Minkyung Cho and {Jy-yong} Sohn},
      year={2024},
      eprint={2412.03223},
      archivePrefix={arXiv},
      primaryClass={cs.CL},
      url={https://arxiv.org/abs/2412.03223}, 
}

@misc{reimers2019sentencebertsentenceembeddingsusing,
      title={Sentence-BERT: Sentence Embeddings using Siamese BERT-Networks}, 
      author={Nils Reimers and Iryna Gurevych},
      year={2019},
      eprint={1908.10084},
      archivePrefix={arXiv},
      primaryClass={cs.CL},
      url={https://arxiv.org/abs/1908.10084}, 
}

@misc{openai_api_reference,
  title        = {OpenAI API Reference},
  author       = {{OpenAI}},
  howpublished = {\url{https://platform.openai.com/docs/api-reference}},
}

@String{Computing = "Computing" }

@ArtifactSoftware{R,
    title = {R: A Language and Environment for Statistical Computing},
    author = {{R Core Team}},
    organization = {R Foundation for Statistical Computing},
    address = {Vienna, Austria},
    year = {2019},
    url = {https://www.R-project.org/},
}


\appendix

\lstset{
  basicstyle=\ttfamily\small,
  breaklines=true,
  frame=single,
  columns=fullflexible
}

\begin{figure*}
\centering
\begin{lstlisting}
# Main Instruction
You are given a single SEC filing document. Your task is to identify and summarize all relevant content in the SEC filing according to the following five categories. Extract and summarize only the information that is relevant to each category.

# Descriptions of the 5 Categories
1. overviewProduct: Provides a high-level summary of the company's business, including its mission, core products or services, key customer groups, business segments, and primary geographic markets.
2. strategyMarketOps: Describes the company's strategic direction and competitive strengths (e.g., proprietary technology, intellectual property, and regulatory expertise), along with its target markets, regulatory context, and operating model such as manufacturing footprint, supply chain structure, and major partnerships.
3. governanceRisks: Covers the company's governance framework, including notable changes in leadership or the board, and summarizes key risks disclosed in filings. Risk factors are organized by category (e.g., regulatory, market, operational, and cybersecurity) and may include both long-term structural risks and near-term concerns.
4. financialStatement: Summarizes the company's financial statements with key figures, and provides an interpretation of its financial health based on filings-discussing major performance drivers, liquidity, funding and capital resources, capital allocation decisions, accounting updates, and material obligations.
5. recentEventCatalyst: Highlights significant developments from roughly the past 12 months, including changes to earnings outlook, major product releases, regulatory decisions, M\&A progress, leadership updates, and other events that could meaningfully affect market perception or performance.

# Rules
1. The SEC filing can be long and may contain noisy formatting. Focus on the content rather than the format.
2. Do not hallucinate. Only use information explicitly stated in the SEC filing.
3. If there is no relevant information for a category, return an empty string for that category.

# Output Format
1. Example: {
  "overviewProduct": ...,
  "strategyMarketOps": ...,
  "financialStatement": ...,
  "governanceRisks": ...,
  "recentEventCatalyst": ...
}

# SEC Filing
[sec_filing_text]
\end{lstlisting}
\vspace{-1em}
\caption{Prompt for LLM-based SEC filing parser.}
\label{fig:prompt_filing_parser}
\vspace{0.5em}
\end{figure*}

\begin{figure*}[t]
\centering
\begin{lstlisting}
# Main Instruction
You are given a single news article. Your task is to analyze the article and classify it into one of the following categories.

# Descriptions of Categories
1. [CATEGORY 1]: ...
2. [CATEGORY 2]: ...
3. [CATEGORY 3]: ...
4. N/A: The article does not fit any of the above categories.

# COMPANY_PROFILE (provided only for company-level classification into Target vs. Related Companies)
1. Financial Statement: [FINANCIAL_STATEMENT]
2. Governance Risks: [GOVERNANCE_RISKS]
3. Overview Product: [OVERVIEW_PRODUCT]
4. Recent Event Catalyst: [RECENT_EVENT_CATALYST]
5. Strategy Market Ops: [STRATEGY_MARKET_OPS]

# Rules
1. Please ignore any unusual or inconsistent formatting in the article.

# Output
1. Example: {"category": ...}
2. "category" must be one of [CATEGORY 1], [CATEGORY 2], [CATEGORY 3], or N/A.

# Article
Headline: [HEADLINE]
Body: [BODY]
\end{lstlisting}
\vspace{-1em}
\caption{Prompt for LLM-based news classification.}
\label{fig:prompt_llm_classification}
\vspace{0.7em}
\end{figure*}

\begin{figure*}
\begin{lstlisting}
# Main Instruction
You will be given a list of multiple ARTICLES that may impact [TAG].
Your task is to review all ARTICLES and identify up to [N] key categories of significant events that may impact [TAG].
For each category, summarize the key factual events in the related ARTICLES.

# Descriptions of [TAG]
[TAG DESCRIPTION]

# Rules
1. Please ignore any unusual or inconsistent formatting in the ARTICLES.
2. Select a category only if it has a meaningful impact on [TAG].
3. If more than [N] categories are identified, select only the [N] most important ones.
4. Each category must address one single, distinct topic only.
5. Do not hallucinate. Write only based on the given ARTICLES.

# Output Format
1. Example: {
    "category1": ...,
    "category2": ...,
    ...
}

# Articles
1. Article 1:
   1-1. Headline: [HEADLINE_1]
   1-2. Body: [BODY_1]
...


\end{lstlisting}
\vspace{-1em}
\caption{Prompt for LLM-based news summarization.}
\label{fig:prompt_llm_summary}
\vspace{2em}                                                                                  
\noindent\hfill{\small\normalfont Received 8 February 2026; accepted 16 May 2026}            
\end{figure*}

\section{LLM Prompts Used in Our Framework}\label{appen:llm_prompt}
In this section, we present the LLM prompts used in our framework. Figure~\ref{fig:prompt_filing_parser} shows the prompt for the LLM-based SEC filing parser, Figure~\ref{fig:prompt_llm_classification} presents the prompt for LLM-based news classification, and Figure~\ref{fig:prompt_llm_summary} illustrates the prompt for LLM-based news summarization. 

In our news classification pipeline, we consider three classification tasks: \textit{i}) categorizing all news into macro-, sector-, and company-level news, \textit{ii}) further classifying sector-level news into 11 GICS sectors, and \textit{iii}) dividing company-level news into related-company and target-company news. For each task, the list of categories corresponding to the placeholder [CATEGORY], along with the meaning of each category, is described as follows.



\begin{enumerate}[leftmargin=*, itemsep=2pt]
    \item \textbf{Macro/Sector/Company-level Classification.}
    \begin{itemize}[leftmargin=0pt, itemsep=1pt, topsep=1pt]
        \item \textbf{Macroeconomic:} News that analyzes or affects the overall U.S. economy.
        \item \textbf{Sector:} News that does not impact the entire U.S. economy, but affects an entire GICS sector or multiple companies within a specific sector.
        \item \textbf{Company:} News that primarily affects a specific company or a small number of individual stocks, rather than the broader economy or a sector.
    \end{itemize}

\item \textbf{Sector Classification: 11 GICS Sectors.}
\begin{itemize}[leftmargin=0pt, itemsep=1pt, topsep=1pt]
    \item \textbf{Energy:} Companies related to oil, gas, coal, and energy services.
    
    \item \textbf{Materials:} Companies producing raw materials such as metals, chemicals, and construction materials.
    
    \item \textbf{Industrials:} Firms involved in manufacturing, transportation, aerospace, construction, and logistics.
    
    \item \textbf{Consumer Discretionary:} Non-essential consumer goods and services such as automobiles, retail, and leisure.
    
    \item \textbf{Consumer Staples:} Essential everyday products including food, beverages, and household goods.
    
    \item \textbf{Health Care:} Pharmaceutical, biotechnology, medical device, and health care service companies.
    
    \item \textbf{Financials:} Banks, insurance companies, asset managers, and financial service providers.
    
    \item \textbf{Information Technology:} Software, semiconductors, hardware, cloud computing, and AI-related companies.
    
    \item \textbf{Communication Services:} Telecommunications, media, entertainment, and internet platform companies.
    
    \item \textbf{Utilities:} Providers of electricity, gas, water, and other public utility services.
    
    \item \textbf{Real Estate:} Real estate developers, property managers, and REITs.
\end{itemize}

    \item \textbf{Company Classification: Target vs. Related Company.}
    \begin{itemize}[leftmargin=10pt, itemsep=1pt, topsep=1pt]
        \item \textbf{Target-company news:} News describing events directly related to the target company’s operations, decisions, financial activities, or other company-specific developments.
        
        \item \textbf{Related-company news:} News describing external events involving competitors, partners, suppliers, or other related entities that may indirectly influence the target company.
    \end{itemize}

\end{enumerate}

In Figure~\ref{fig:prompt_llm_summary}, the placeholder [TAG] corresponds to the final classification label, which can be one of Macroeconomic, one of the 11 GICS sector names, or a specific company name. For Macroeconomic and the 11 GICS sector tags, the corresponding category description is provided as [TAG DESCRIPTION]. For company-level tags, [TAG DESCRIPTION] is populated with  \textit{overviewProduct}, \textit{strategyMarketOps}, \textit{governanceRisks}, \textit{financialStatement}, and \textit{recentEventCatalyst}.

\makeatletter
\global\let\@received\@empty   
\makeatother
\end{document}